\documentclass{article}

\usepackage{arxiv}

\usepackage[utf8]{inputenc}
\usepackage[T1]{fontenc}
\usepackage{hyperref}
\usepackage{url}
\usepackage{booktabs}
\usepackage{amsfonts}
\usepackage{amsmath}
\usepackage{nicefrac}
\usepackage{microtype}
\usepackage{xcolor}
\usepackage{graphicx}
\usepackage{subcaption}
\usepackage{natbib}

\title{MindGames Arena Generalization Track:\\In2AI Solution with Delayed Per-Step Reward Attribution}

\author{
 Aliaksei Korshuk \\
 iMak AI Lab \\
 Coframe \\
 \texttt{a.korshuk@innopolis.university} \\
 \And
 Alexander Buyantuev \\
 iMak AI Lab \\
 \And
 Ilya Makarov \\
 iMak AI Lab \\
}

\renewcommand{\headeright}{Technical Report}
\renewcommand{\undertitle}{Technical Report}
\renewcommand{\shorttitle}{MindGames Arena Generalization Track: In2AI Solution}

\hypersetup{
pdftitle={MindGames Arena Generalization Track: In2AI Solution with Delayed Per-Step Reward Attribution},
pdfauthor={Aliaksei Korshuk, Alexander Buyantuev, Ilya Makarov},
pdfkeywords={Reinforcement Learning, Multi-Agent Systems, Language Models, Reward Attribution},
}

\begin{document}

\maketitle

\begin{abstract}
Training language model agents for multi-agent strategic interaction presents a core difficulty: the quality of any action may depend on future events that never materialize, on moves that violate game rules, or on decisions made by other players. Standard reinforcement learning assumes that rewards can be assigned at each step, but this assumption fails in settings where outcomes are entangled across time and agents. We introduce \emph{delayed per-step reward attribution with eligibility gating}, an episode lifecycle and postprocessing pipeline that computes rewards only at episode end, propagates them back to originating steps according to task-specific semantics, and excludes steps that lack valid dependent information from training. Together with asynchronous rollout generation via vLLM's continuous batching, curriculum-based opponent sampling, and multi-level stratified batch construction, this approach enables stable, sample-efficient RL training in multi-agent environments. We evaluate on the MindGames Arena benchmark at NeurIPS 2025, where a single 8-billion-parameter open-source model trained with our method matched or surpassed substantially larger proprietary systems, including GPT-5, in head-to-head play and took first place in both the Open (unrestricted) and Efficient ($\leq$8B parameters) tracks.
\end{abstract}


\section{Introduction}
\label{sec:introduction}

Large language models excel at single-turn tasks \citep{deepseekai2025deepseekr1incentivizingreasoningcapability}, yet settings that require modeling other agents' beliefs, coordinating under uncertainty, and planning over extended interactions remain difficult \citep{guertler2025textarena, park2023generative}. Recent work applies reinforcement learning to improve LLM agents \citep{ouyang2022instructgpt, shinn2023reflexion}, but most methods assume single-agent, single-turn environments where reward signals are immediate and well-defined.

Multi-agent strategic interaction fundamentally violates this assumption. When language model agents interact over extended episodes, whether negotiating, deceiving, cooperating, or competing, \textbf{the quality of an action depends on events beyond the agent's control}. A negotiation opening is good or bad depending on how the counterparty responds; a clue is clever or foolish depending on whether the teammate interprets it correctly; a deceptive allocation is brilliant or wasteful depending on whether the opponent takes the bait. Traditional RL frameworks that assign immediate rewards cannot capture these dependencies.

We introduce \emph{delayed per-step reward attribution with eligibility gating}: rather than forcing immediate reward assignment, we (1) determine \emph{at episode end} who won and compute episode-level rewards, (2) \emph{gate} steps that lack sufficient training signal, and (3) attribute rewards \emph{backward} according to task-specific semantics. Paired with asynchronous rollout generation, curriculum-based opponent sampling, and multi-level stratified batch construction, this approach yields stable reinforcement learning in multi-agent settings. We evaluate on the MindGames Arena benchmark \citep{mindgames2025}, powered by the TextArena framework \citep{guertler2025textarena}, where our 8B-parameter model placed first in \emph{both} the Open (unrestricted) and Efficient ($\leq$8B) divisions, matching or beating teams that used GPT-5 \citep{openai2025gpt5} in head-to-head play.

\section{Related Work}
\label{sec:related}

\paragraph{Reinforcement learning for language models.}
Reinforcement learning from human feedback (RLHF) is now the dominant method for aligning LLMs with human preferences \citep{ouyang2022instructgpt}, typically using proximal policy optimization \citep{schulman2017ppo} or REINFORCE-style methods \citep{ahmadian2024basicsrevisitingreinforcestyle}. These methods operate in a single-turn, single-agent setting where reward is immediate and well-defined. Recent work extends RL to multi-turn settings: \citet{shinn2023reflexion} introduce verbal self-reflection as a reinforcement signal, and \citet{snell2024testtime} study how scaling test-time compute can improve LLM reasoning, sometimes more effectively than scaling model parameters. DeepSeek-R1 \citep{deepseekai2025deepseekr1incentivizingreasoningcapability} demonstrates that pure RL can produce emergent chain-of-thought reasoning without supervised warm-up, but focuses on single-agent mathematical reasoning rather than multi-agent interaction.

\paragraph{Multi-agent reinforcement learning.}
Training strategically interacting agents is a well-studied problem in multi-agent RL \citep{zhang2021multiagent}. Centralized training with decentralized execution (CTDE) methods such as MAPPO \citep{yu2022mappo} and QMIX \citep{rashid2020qmix} address credit assignment in cooperative settings, but assume access to shared state information unavailable in natural language games. The credit assignment problem (determining which agent's action caused a collective outcome) is closely related to our work, though we operate at the level of per-step reward attribution within a single agent's trajectory rather than across agents.

\paragraph{LLM agents for games and strategic reasoning.}
TextArena \citep{guertler2025textarena} provides a framework for evaluating LLMs in text-based multi-agent games, and the MindGames Arena \citep{mindgames2025} builds on it to benchmark strategic reasoning across cooperative, competitive, and mixed-motive settings. SPIN-Bench \citep{yao2025spinbench} evaluates strategic planning, interaction, and negotiation capabilities. Prior work on game-playing LLMs has largely focused on prompting strategies \citep{gandhi2023strategic} or few-shot evaluation \citep{akata2023playing} rather than training agents through RL in game environments. Axelrod's foundational work on the evolution of cooperation \citep{axelrod1981evolution} and Blotto games \citep{roberson2006colonel} provide the game-theoretic basis for two of our evaluation environments.

\paragraph{Curriculum learning and opponent diversity.}
Self-play and population-based training \citep{jaderberg2019pbt} work well for training game-playing agents but typically operate in fixed-action-space games. Our curriculum approach, which gradually introduces stronger opponents while retaining weaker ones, draws on ideas from prioritized experience replay \citep{schaul2016prioritized} and automated curriculum learning \citep{portelas2020automatic}, adapted to the LLM setting where opponent diversity shapes the distribution of natural language strategies encountered during training.

\section{Game Environments}
\label{sec:competition}

MindGames Arena \citep{mindgames2025} evaluates whether language model agents can reason strategically, coordinate with partners, and adapt to opponents while communicating in natural language. The Generalization Track includes two divisions:
\begin{itemize}
 \item \textbf{Efficient:} Open-source models with at most 8 billion parameters.
    \item \textbf{Open:} Any models without constraints on size or cost.
\end{itemize}

Agents play many matches against varied opponents, with results aggregated using TrueSkill \citep{herbrich2006trueskill} to produce reliable rankings.

The benchmark includes three games that span cooperative, competitive, and mixed-motive dynamics:

\paragraph{Codenames.} A cooperative 2v2 word-association game. A 5$\times$5 grid contains 25 words: 9 Red team words, 8 Blue team words, 7 neutral words, and 1 assassin. Each team has a \emph{spymaster} (who sees word assignments) and an \emph{operative} (who only sees revealed words). The spymaster gives a one-word clue plus a number indicating how many team words relate to that clue (format: \texttt{[clue N]}). The operative then guesses up to $N+1$ words sequentially (format: \texttt{[word]} or \texttt{[pass]}). Guessing the assassin causes instant loss; guessing opponent or neutral words ends the turn. The first team to reveal all their words wins; if the 80-turn limit is reached, the team with more revealed words wins.

\paragraph{Colonel Blotto.} A two-player competitive resource allocation game. Each round, both players simultaneously allocate exactly 20 units across 3 battlefields (format: \texttt{[A5 B10 C5]}). The player committing more units to a field wins that field; winning the majority of fields wins the round. The match continues for up to 9 rounds, with early victory possible upon securing a majority (5+). The player winning more rounds wins the match.

\paragraph{Three-Player Iterated Prisoner's Dilemma.} A mixed-motive game extending the classic Iterated Prisoner's Dilemma to three players over 5 rounds. Each round has two phases: (1) \emph{conversation}, consisting of 3 free-chat turns where players negotiate openly, and (2) \emph{decision}, where each player simultaneously submits actions toward both opponents (format: \texttt{[1 cooperate] [2 defect]}). The payoff matrix follows the standard structure: mutual cooperation yields 3 points each; mutual defection yields 1 point each; unilateral defection yields 5 points (defector) and 0 points (cooperator). The player(s) with the highest cumulative score after all rounds wins.

\section{Challenges in Agentic Workflows}
\label{sec:challenges}

Standard reinforcement learning assumes a convenient fiction: that each action can be evaluated independently, with rewards fully capturing the quality of that action at the moment it is taken. This assumption underlies everything from temporal difference updates in Q-learning \citep{watkins1992qlearning} to per-step advantage estimates in policy gradient methods \citep{schulman2017ppo}. In single-agent, fully-observable environments with dense rewards, this fiction holds reasonably well.

Agentic workflows break this assumption. When language model agents interact over extended episodes, the quality of an action becomes \emph{entangled} with events that have not yet occurred, with actions taken by other agents, and with information that may never be revealed.

Before presenting our solution, we describe eight specific challenges that motivated our design. These challenges extend well beyond MindGames \citep{mindgames2025}; they arise in any agentic workflow where action outcomes depend on future events, other agents, or information outside the actor's control. We group them into three themes: \emph{temporal entanglement} (Challenges 1, 2), where action quality depends on future events; \emph{structural asymmetry} (Challenges 3, 4, 5), where position, opponent skill, or external failures create unequal learning signals; and \emph{training logistics} (Challenges 6, 7, 8), where variable episode structure and heterogeneous inference demands complicate batch construction and throughput.

\subsection{Challenge 1: Lose to Win}
\label{sec:challenge-lose}

Consider the following Colonel Blotto exchange:

\begin{center}
\small
\begin{tabular}{lcllc}
\toprule
\textbf{Round} & \textbf{Alpha} & \textbf{Beta} & \textbf{Result} & \textbf{Post-Round Score (Alpha vs Beta)} \\
\midrule
7 & \texttt{[A3 B8 C9]} & \texttt{[A0 B10 C10]} & Beta wins & 2 vs 4 \\
8 & \texttt{[A1 B11 C8]} & \texttt{[A0 B10 C10]} & Alpha wins & 3 vs 4 \\
\multicolumn{5}{l}{\textit{Alpha observes: ``Beta repeats the same allocation. I'll exploit this pattern.''}} \\
9 & \texttt{[A1 B11 C8]} & \texttt{[A2 B12 C6]} & Beta wins & 3 vs 5 \\
\midrule
\multicolumn{4}{l}{\textbf{Match Result:}} & \textbf{Beta wins} \\
\bottomrule
\end{tabular}
\end{center}

Beta's Round 8 allocation, \texttt{[A0 B10 C10]}, identical to Round 7, \emph{lost} that round. Yet this move was strategically brilliant: it established a predictable pattern that Alpha learned to exploit. When Beta broke the pattern in Round 9, Alpha's counter-strategy backfired.

\textbf{Problem:} Locally suboptimal actions may be globally optimal. A reward function that evaluates each step on its immediate outcome would penalize setup actions that sacrifice short-term performance for long-term success, discouraging strategies such as deception, baiting, or delayed gratification.

\textbf{Implication:} Rewards must incorporate final outcome and propagate backward to preceding actions. Intermediate results alone are insufficient to evaluate action quality.

\subsection{Challenge 2: Interdependent Action Rewards}
\label{sec:challenge-undefined}

Consider this Codenames episode where the team words include ``ocean'' and ``fish'':

\begin{quote}
\small
\textbf{Spymaster} gives clue: \texttt{[water 2]}

\textbf{Operative} reasons: \textit{``Options: ocean, shower, blue, fish. With 2 guesses, I'll try ocean and fish...''}

\textbf{Operative} guesses: \texttt{[ocean]} \hfill $\checkmark$ \textit{Correct! Team word revealed.}

\textbf{Operative} reasons: \textit{``Ocean was right. Remaining: shower, blue, fish. Since ocean is blue, maybe `blue' is the connection...''}

\textbf{Operative} guesses: \texttt{[blue]} \hfill $\times$ \textit{Assassin! Game over.}
\end{quote}

Despite being correct, the first guess \texttt{[ocean]} should not receive maximum reward: the turn ended in catastrophe, and all guesses and the clue share responsibility.

\textbf{Problem:} Actions within a logical sequence are interdependent. Early actions depend on \emph{future} actions for their true value; later actions inherit context from \emph{past} actions; initiating actions depend on \emph{all} subsequent responses they trigger. Per-step rewards that evaluate actions independently reinforce locally correct actions even when they contribute to global failure.

\textbf{Implication:} All actions within a logical group must be evaluated together, with rewards computed after observing the complete sequence and attributed based on collective outcome.

\subsection{Challenge 3: Positional Bias}
\label{sec:challenge-position}

Consider the Three-Player Iterated Prisoner's Dilemma conversation phase, where players speak in fixed order (1 $\rightarrow$ 2 $\rightarrow$ 3) before submitting simultaneous decisions:

\begin{itemize}
    \item \textbf{Player 1 (early position):} Speaks first, can propose coalitions (e.g., ``Let's both defect against Player 3''), and observes reactions before the decision phase.
    \item \textbf{Player 2 (middle position):} Can respond to prior proposals and add their own, receiving partial feedback.
    \item \textbf{Player 3 (late position):} Speaks last. Any proposals they make are \emph{never discussed} before decisions are locked in.
\end{itemize}

\textbf{Problem:} Turn order confers asymmetric strategic affordances. Early actors set agendas and observe responses; late actors must interpret prior commitments without any chance to clarify. An agent trained mostly from one position learns position-specific strategies that do not transfer to other positions.

\textbf{Implication:} Training must expose the agent to all positions uniformly to learn position-agnostic strategies. Skewed sampling produces brittle policies that break under position reassignment.

\subsection{Challenge 4: Opponent and Teammate Diversity}
\label{sec:challenge-curriculum}

A model that has not yet learned game rules and action formats will produce invalid outputs. If this untrained model immediately faces frontier opponents like GPT-5 \citep{openai2025gpt5}, every episode ends in rapid defeat before the model can observe what correct play looks like. The training signal becomes dominated by format penalties rather than strategic learning.

Conversely, training exclusively against a narrow set of opponents or teammates produces a model that exploits their specific patterns but fails against diverse real-world agents. In cooperative games like Codenames, a spymaster trained only with one type of operative learns to give clues tailored to that partner's interpretation style, failing when paired with different teammates at evaluation time.

\textbf{Problem:} The training distribution of opponents and teammates must be designed along two axes. First, \emph{skill progression}: opponents that are too strong yield no positive signal for a novice agent, while opponents that are too weak fail to challenge an improving one. Second, \emph{diversity}: training against a homogeneous set of agents produces brittle policies that overfit to specific play styles instead of learning strategies that generalize.

\textbf{Implication:} Agent sampling must provide both curriculum learning and diversity. Training should start with opponents that let the model learn basic rules and formats, then gradually introduce stronger and more varied opponents as skill improves. Earlier opponents must remain in the sampling pool throughout training, both to prevent catastrophic forgetting and to ensure the model continues to beat weaker opponents while learning to compete with stronger ones.

\subsection{Challenge 5: Missing Training Signal}
\label{sec:challenge-missing}

Consider two failure scenarios:

\paragraph{Colonel Blotto.} Player Alpha submits a valid allocation \texttt{[A5 B10 C5]}, but opponent Beta submits \texttt{[A100 B0 C0]} (invalid: sum exceeds 20). The match terminates immediately. There was no battlefield comparison, no round outcome. The action was valid, but there is no signal about whether it was strategically sound.

\paragraph{Codenames.} The spymaster gives a valid clue \texttt{[water 2]}. The operative responds with ``My answer is ocean'' instead of the required format \texttt{[ocean]}. The turn terminates due to the parsing error. The clue has no guesses to evaluate, making it impossible to determine its quality.

\textbf{Problem:} Some valid actions lack observable outcomes due to external failures (opponent errors, parsing failures, early termination). Assigning arbitrary rewards to these steps introduces noise; assigning zero rewards biases against exploratory actions. Neither approach provides a meaningful learning signal.

\textbf{Implication:} Steps without sufficient signal must be identified and excluded from training. Only actions with observable outcomes should contribute to gradient updates.

\subsection{Challenge 6: Variable Episode Length}
\label{sec:challenge-variable}

Agentic episodes exhibit high variance in length. A game may terminate early through decisive victory, end abruptly due to an invalid action, extend through prolonged negotiation toward a draw, or reach a predefined step limit. In Codenames, hitting the assassin terminates immediately, while revealing all words through careful play may take substantially longer. Table~\ref{tab:episode_length} illustrates this variability across our training data.

\begin{table}[h]
\centering
\caption{Episode trainable length statistics across game environments. \emph{Avg Steps}: mean number of trainable agent steps per episode. \emph{Std}: standard deviation. \emph{Games}: number of episodes sampled. \emph{Min/Max}: shortest and longest observed number of trainable steps.}
\label{tab:episode_length}
\begin{tabular}{lccccc}
\toprule
Environment & Avg Steps Per Player & Std & Games & Min & Max \\
\midrule
Codenames & 5.18 & 4.30 & 56 & 1 & 16 \\
Colonel Blotto & 6.48 & 1.61 & 426 & 1 & 9 \\
Three-Player IPD & 9.77 & 1.18 & 262 & 2 & 10 \\
\bottomrule
\end{tabular}
\end{table}

This variability compounds with eligibility filtering: not all steps in an episode qualify for training. An episode with 15 steps may yield only 8 eligible training samples after removing invalid actions and steps that lack signal. Rollout generation must collect episodes into an episodes bank, accumulating eligible steps until enough samples are available for a training batch.

\textbf{Problem:} Standard reinforcement learning pipelines assume fixed or predictable episode lengths for batch construction. Variable-length episodes with dynamic eligibility create uneven data flow: some rollouts produce many training samples, others produce few or none. Synchronous collection idles compute while waiting for long episodes; fixed batch sizes waste capacity or train on stale data.

\textbf{Implication:} Rollout generation must run asynchronously across parallel workers and respect the actual number of trainable steps each episode provides when assembling training batches.

\subsection{Challenge 7: Multi-Dimensional Batch Balancing}
\label{sec:challenge-balancing}

As Table~\ref{tab:episode_length} illustrates, different games produce episodes of vastly different lengths. When training a single model across multiple games, unbalanced sampling causes the model to see disproportionately more steps from longer-episode games, biasing learned policies toward those environments.

Even within a single game, episode length varies substantially (note the standard deviations in Table~\ref{tab:episode_length}). An episodes bank containing a mix of long and short episodes may have most of its steps concentrated in a few long trajectories. Sampling uniformly by step would draw disproportionately from these long episodes, reducing diversity: the model repeatedly sees correlated steps from the same trajectory rather than learning from varied situations across many episodes.

Reward distributions are also skewed. Most steps cluster around average rewards, while highly positive (excellent moves) and highly negative (critical errors) steps are rarer. Uniform sampling underrepresents these extremes, yet the model must see failures to avoid them and successes to reinforce them.

\textbf{Problem:} Uniform step sampling produces imbalanced coverage across three dimensions: games, episodes, and reward outcomes. Long-episode games dominate gradients; long individual episodes reduce batch diversity; skewed reward distributions cause the model to underlearn from rare but important successes and failures.

\textbf{Implication:} Batch construction must balance across games, episodes, and reward bins. Steps should be sampled for even game representation, broad episode coverage, and stratified reward sampling that covers bad, average, and good actions proportionally.

\subsection{Challenge 8: Heterogeneous Inference Demands}
\label{sec:challenge-inference}

Different game situations demand vastly different amounts of reasoning. A spymaster crafting a clue that connects multiple words while avoiding the assassin requires extensive chain-of-thought deliberation, while an operative selecting a single word from a short list needs far less reasoning. Similarly, negotiating a coalition in the Prisoner's Dilemma requires more deliberation than submitting a binary cooperate/defect decision. These differences appear both across games and across roles within the same game.

When multiple workers run episodes concurrently (as required by Challenge~\ref{sec:challenge-variable}), these heterogeneous demands create synchronization bottlenecks. Standard inference pipelines process requests in synchronized batches: all requests in a batch must complete before any results are returned. A worker generating a short response cannot proceed while another worker in the same batch deliberates extensively.

\textbf{Problem:} Synchronous batch inference creates idle time proportional to the variance in generation length. If $N$ workers submit requests with generation times $t_1, \ldots, t_N$, synchronous batching forces all workers to wait $\max_i(t_i)$ before any can proceed, wasting $\sum_i (\max_j(t_j) - t_i)$ worker-seconds per inference round. With high variance across games and roles, this overhead compounds multiplicatively across episodes, dominating training time and reducing throughput.

\textbf{Implication:} The inference engine must support asynchronous request handling with continuous batching: requests should be processed as they arrive, with results returned upon individual completion rather than held for batch synchronization. Each worker must be able to proceed independently so that episodes advance at their natural pace.

\subsection{Summary: A Unified Problem}
\label{sec:challenge-summary}

These eight challenges are not independent obstacles to be addressed in isolation; they form a \emph{coupled problem} stemming from the mismatch between traditional RL assumptions and the realities of agentic interaction.

The temporal entanglement challenges (Lose to Win, Interdependent Rewards) show that \textbf{reward computation must be delayed}: we cannot assign meaningful rewards until we observe how actions interact with future events. The structural asymmetry challenges (Positional Bias, Opponent/Teammate Diversity, Missing Training Signal) show that \textbf{not all training configurations are equal}: some lack valid signal entirely, others encode position-specific strategies that do not generalize, and still others arise from homogeneous opponent distributions that produce brittle policies. The logistical challenges (Variable Episode Length, Multi-Dimensional Balancing, Heterogeneous Inference) show that \textbf{episode structure and inference demands are unpredictable}: training pipelines must tolerate high variance in episode length, step eligibility, reward distribution, and per-step generation time.

Attempting to solve any single challenge in isolation creates new problems. Delaying rewards without filtering steps assigns arbitrary values to undefined outcomes. Filtering steps without balanced sampling biases the model toward easy cases. Balanced sampling without proper reward attribution trains on noisy gradients. Asynchronous rollout generation without continuous-batching inference introduces synchronization bottlenecks. \textbf{The challenges compound}, and no piecemeal fix suffices; the system must address all three themes jointly.

\section{Methodology}
\label{sec:approach}

We address the challenges outlined in Section~\ref{sec:challenges} through an \emph{episode lifecycle design} that separates action validation during execution from reward computation after episode completion. The central idea is that reward assignment must be \emph{delayed} until sufficient information is available, and steps that lack valid training signal must be \emph{filtered} from the training batch rather than assigned arbitrary rewards.

Our approach comprises two components: (1) \emph{action validation} that enforces format and rule compliance during episode execution, and (2) a \emph{post-episode processing pipeline} that computes rewards, gates eligibility, and attributes credit based on complete trajectory information. We describe each in turn.

\subsection{Action Validation During Execution}

The TextArena framework \citep{guertler2025textarena} does not uniformly terminate episodes on invalid actions: some games continue with default behaviors, others ignore malformed outputs, and error handling varies across environments. There is also no unified interface for determining whether an action was incorrect: each environment reports errors differently, if at all. This inconsistency complicates reward attribution because we cannot reliably determine which steps failed or why.

To address this, we introduce an \textbf{Action Validator} abstraction for each environment. During episode execution, every action passes through the validator, which performs three checks:
\begin{itemize}
    \item \textbf{Reasoning template compliance:} Does the output follow the expected reasoning structure? (e.g., thinking inside designated tags before the final answer)
    \item \textbf{Format compliance:} Does the output match the expected action pattern? (e.g., \texttt{[word N]} for spymaster clues, \texttt{[A5 B10 C5]} for Blotto allocations)
    \item \textbf{Game-rule validity:} Does the action satisfy game constraints? (e.g., allocation sum $\leq 20$ in Blotto, clue not being a substring of board words in Codenames, guess being an existing word on the board)
\end{itemize}

Invalid actions terminate the episode immediately with a recorded error type. This uniform validation serves several downstream purposes: it provides the metadata needed to identify \emph{Missing Training Signal} (Challenge~\ref{sec:challenge-missing}) by detecting when valid actions lack outcomes due to external failures; it establishes clear episode boundaries for \emph{delayed reward attribution} (Challenges~\ref{sec:challenge-lose} and~\ref{sec:challenge-undefined}); and it lets us reliably determine who won, lost, or caused premature termination. The Players Builder (Section~\ref{sec:players-builder}) relies on this metadata to extract episode-level results, the Steps Filter uses it to gate steps that depend on failed actions, and the Reward Assigner applies appropriate penalties to invalid outputs.

\subsection{Post-Episode Processing Pipeline}

\begin{figure}[t]
    \centering
    \includegraphics[width=\textwidth]{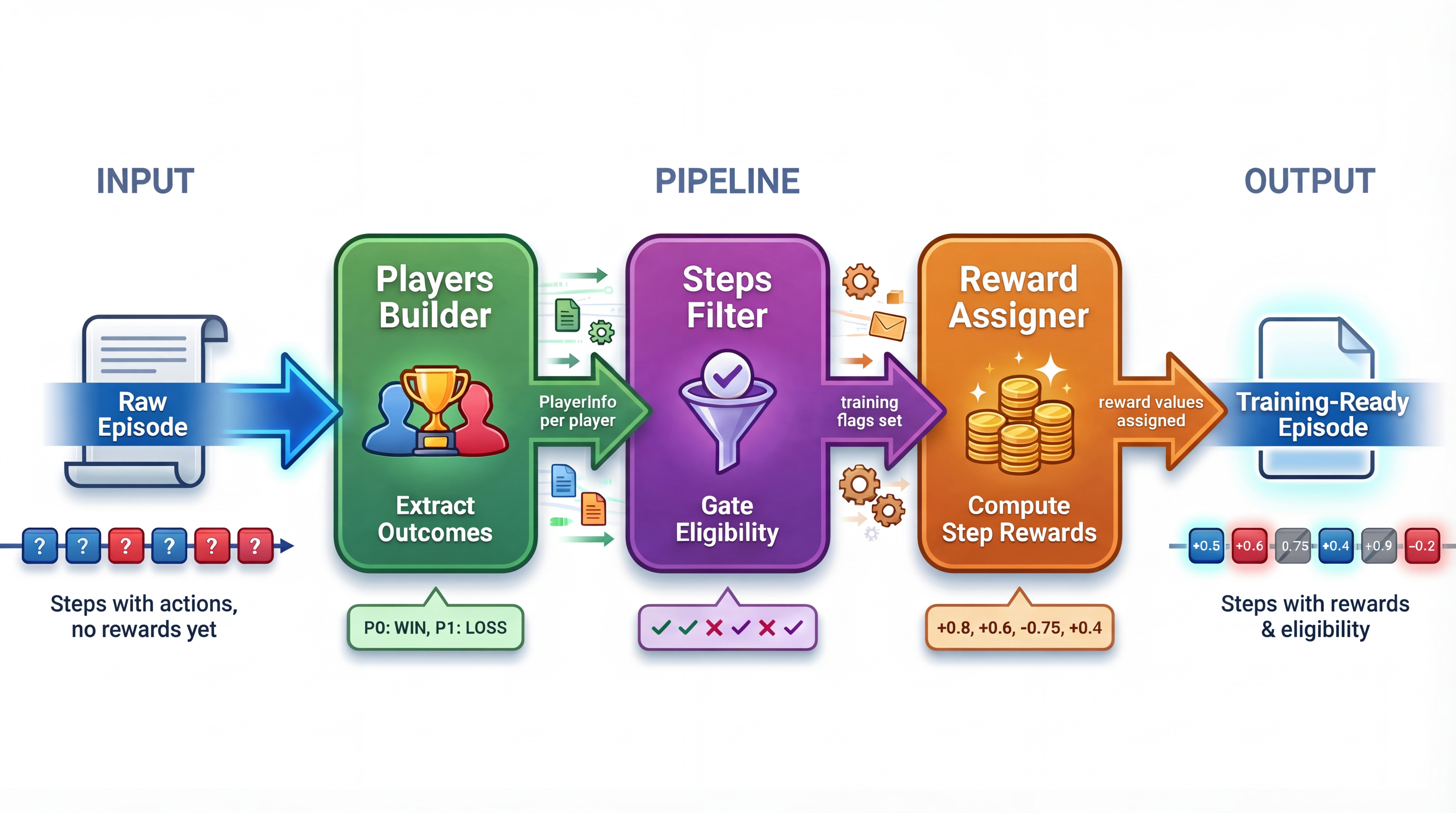}
    \caption{\textbf{Post-episode processing pipeline.} Raw episodes enter with actions but no rewards. The \textbf{Players Builder} extracts outcomes and episode-level rewards. The \textbf{Steps Filter} gates eligibility based on signal availability. The \textbf{Reward Assigner} computes per-step rewards using delayed attribution. Output: training-ready episodes with rewards and eligibility flags.}
    \label{fig:pipeline}
\end{figure}

After episode termination, each completed episode passes through a three-stage processing pipeline (Figure~\ref{fig:pipeline}) that transforms raw trajectories into training-ready data with per-step reward attribution. The \textbf{Players Builder} extracts episode outcomes for delayed evaluation; the \textbf{Steps Filter} gates steps that lack sufficient signal (Challenge~\ref{sec:challenge-missing}); and the \textbf{Reward Assigner} computes per-step rewards with backward propagation (Challenges~\ref{sec:challenge-lose} and~\ref{sec:challenge-undefined}).

\subsubsection{Stage 1: Players Builder}
\label{sec:players-builder}

The \textbf{Players Builder} extracts the episode outcomes that subsequent stages require. Using the action validation metadata, it reconstructs the full episode result: not just win/loss/draw, but also \emph{how} the episode ended (natural conclusion, parsing failure, or game-rule violation).

For each player, we determine:
\begin{itemize}
    \item \textbf{Outcome type:} Win, loss, draw, or termination due to error (parsing failure vs.\ invalid action)
    \item \textbf{Responsibility:} Did this player cause the termination, or was it caused by an opponent or teammate?
    \item \textbf{Episode-level rewards:} Episode rewards returned by TextArena \citep{guertler2025textarena} are nearly boolean, identifying win or loss but providing little granularity. We compute meaningful rewards that better measure model performance. For example, in Colonel Blotto with 9 rounds and majority-win termination, we normalize by rounds won: 1.0 means winning 5 rounds, 0.0 means zero rounds won, with intermediate values reflecting partial success. Rewards are further adjusted based on termination cause: a player whose opponent made an invalid move receives a different reward than one who won through legitimate play.
\end{itemize}

This per-player outcome analysis matters most in multi-agent games where teammates share credit and opponents may cause early termination. The episode-level rewards serve \emph{logging and metrics}; the actual training signal comes from per-step rewards computed by the Reward Assigner (Section~\ref{sec:reward-assigner}). The structured metadata (outcome type, responsibility, episode reward) feeds into subsequent pipeline stages for step-level processing. Complete formulas appear in Appendix~\ref{app:players-builder}.

\subsubsection{Stage 2: Steps Filter (Eligibility Gating)}

The \textbf{Steps Filter} determines which steps contain valid training signal, targeting the \emph{Missing Training Signal} problem (Challenge~\ref{sec:challenge-missing}). A step is marked \emph{ineligible} (gated) if:

\begin{enumerate}
    \item It belongs to a non-trainable player (opponent).
    \item Its reward depends on future steps that are invalid or missing.
    \item It represents an incomplete interaction unit.
\end{enumerate}

The core principle is: \textbf{no observable outcome $\Rightarrow$ no training signal $\Rightarrow$ step is gated}. Rather than assigning arbitrary rewards to steps with undefined outcomes, we simply exclude them from the training batch. Two concrete examples illustrate this:

\paragraph{Example 1: Opponent failure in Colonel Blotto.} Suppose our trainable player (Alpha) submits a valid allocation \texttt{[A5 B10 C5]}, but the opponent (Beta) submits an invalid allocation \texttt{[A100 B0 C0]} (sum exceeds 20). The match terminates immediately due to the error from Beta. The valid action from Alpha has \emph{no outcome to learn from}: there was no battlefield comparison, no round winner, no signal about whether the strategy was good or bad. The step is \textbf{gated}. However, the invalid action from Beta \emph{is} trained on with a penalty reward, teaching error avoidance.

\paragraph{Example 2: Invalid guess format in Codenames.} The spymaster gives a valid clue \texttt{[water 2]}. The operative responds with ``My answer is ocean'' instead of the required format \texttt{[ocean]}. The turn terminates due to the format error. The clue has \emph{no guesses to evaluate}: we cannot determine whether ``water 2'' was a good or bad clue because no valid guesses followed. The clue step is \textbf{gated}. The malformed response from the operative is trained on with a penalty to teach format compliance.

An important distinction: \emph{error steps remain eligible}. If an action fails validation, we train on it with a penalty reward. The gating applies only to steps where \emph{no outcome exists to evaluate}, for instance when a parsing failure prevents any valid response. Bad outcomes (e.g., hitting the assassin) are \emph{not} gated; they produce observable results that inform reward computation. In such cases, the Reward Assigner (Section~\ref{sec:reward-assigner}) handles blame attribution and may propagate penalties backward to earlier steps such as the originating clue.

\subsubsection{Stage 3: Reward Assigner (Delayed Attribution)}
\label{sec:reward-assigner}

TextArena \citep{guertler2025textarena} does not provide meaningful per-step rewards; it returns only episode-level outcomes (e.g., $+1$ for win, $-1$ for loss). To obtain fine-grained credit assignment and address the temporal entanglement challenges (\emph{Lose to Win}, Challenge~\ref{sec:challenge-lose}; \emph{Interdependent Action Rewards}, Challenge~\ref{sec:challenge-undefined}), we define a \textbf{Reward Assigner} for each environment. The Reward Assigner computes per-step rewards from full trajectory information following three principles:

\paragraph{Backward propagation.} The reward for a step may depend on what happens \emph{after} it. Actions within a turn form interdependent groups where later outcomes affect earlier rewards. Consider two Codenames scenarios:

\begin{quote}
\small
\textbf{Success case:} Spymaster gives clue \texttt{[animal 2]}. Operative guesses \texttt{[dog]} $\checkmark$, then \texttt{[cat]} $\checkmark$, then \texttt{[pass]}.

$\rightarrow$ All guesses receive positive rewards; the clue receives credit based on guess accuracy.
\end{quote}

\begin{quote}
\small
\textbf{Assassin case:} Spymaster gives clue \texttt{[danger 2]}. Operative guesses \texttt{[sword]} $\checkmark$, then \texttt{[assassin]} $\times$.

$\rightarrow$ The assassin guess receives a large penalty ($-0.70$). This penalty \textbf{poisons the entire group}: the correct guess \texttt{[sword]} receives reduced reward (+0.15 instead of +1.0), and the clue is \textbf{blamed} for leading to the assassin ($-0.17$).
\end{quote}

This backward propagation ensures that all dependent actions share responsibility for collective outcomes. The spymaster cannot escape blame for a clue that led the operative to the assassin, even if individual guesses were locally reasonable.

\paragraph{Outcome modulation.} Per-step rewards are scaled by episode outcome, targeting the \emph{Lose to Win} problem (Section~\ref{sec:challenge-lose}): actions in winning games receive full credit regardless of intermediate round results, while the same actions in losing games receive reduced credit. The ``losing'' allocation that sets up a winning deception is credited appropriately.

\paragraph{Group-based attribution.} In Iterated Prisoner's Dilemma, conversation turns and the subsequent decision form a ``group.'' The decision outcome determines the reward, which is then distributed to all group members. This addresses the \emph{Interdependent Rewards} challenge (Section~\ref{sec:challenge-undefined}): all actions within a logical sequence are evaluated together based on collective outcome.

Complete reward formulas for each environment are provided in Appendix~\ref{app:rewards}.

\subsection{Training System}
\label{sec:infrastructure}

\begin{figure}[t]
    \centering
    \includegraphics[width=\textwidth]{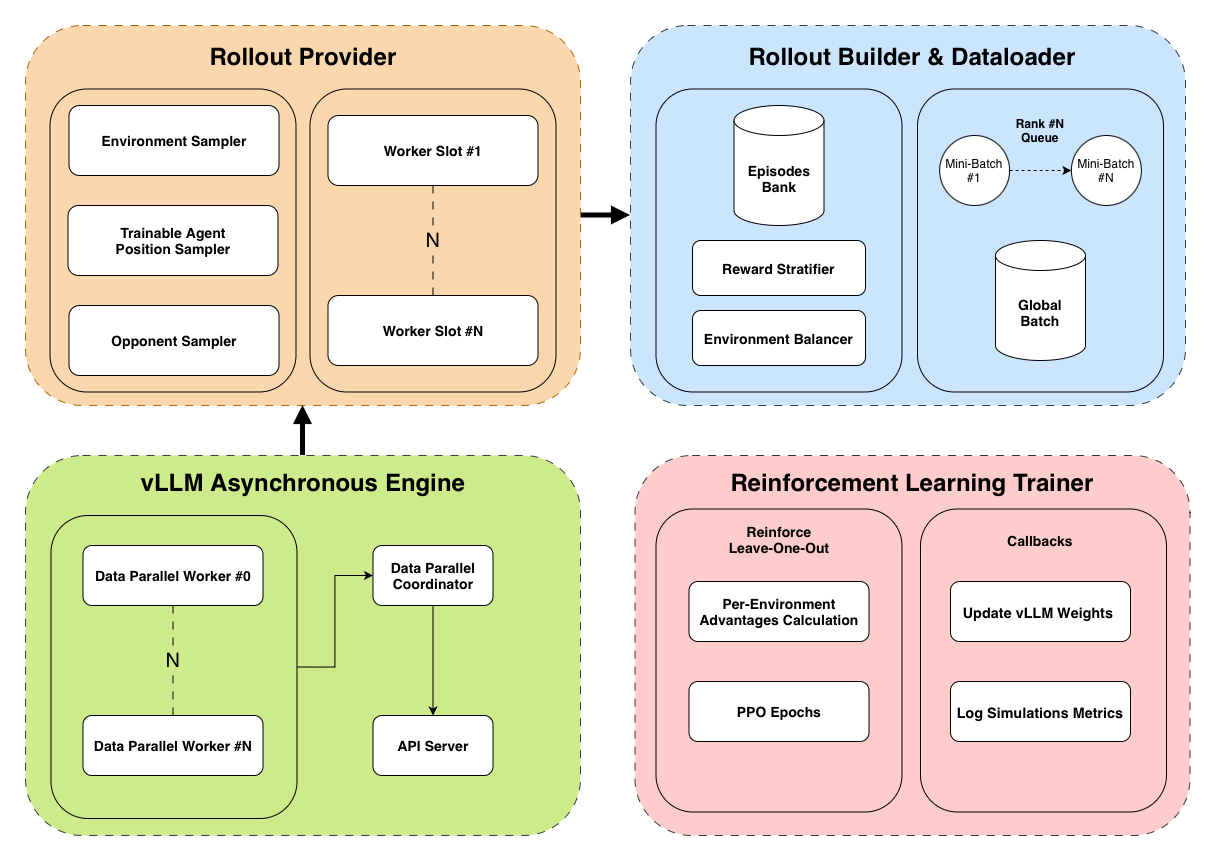}
    \caption{\textbf{System architecture overview.} The \textbf{vLLM Asynchronous Engine} \citep{kwon2023vllm} serves inference requests through an API server with data-parallel workers and continuous batching so that workers need not wait for each other (Challenge~\ref{sec:challenge-inference}). The \textbf{Rollout Provider} queries vLLM to run concurrent episode workers; each slot contains the action validator and post-episode processing pipeline, which computes delayed rewards with backward attribution and filters steps that lack signal (Challenges~\ref{sec:challenge-lose},~\ref{sec:challenge-undefined},~\ref{sec:challenge-missing}); the samplers enforce balanced position coverage and curriculum-based opponent selection (Challenges~\ref{sec:challenge-position},~\ref{sec:challenge-curriculum}). Completed episodes flow to the \textbf{Rollout Builder}, which aggregates them into an episodes bank, applies reward stratification and environment balancing, and constructs per-rank mini-batch queues into a global training batch (Challenges~\ref{sec:challenge-variable},~\ref{sec:challenge-balancing}). The \textbf{Reinforcement Learning Trainer} optimizes the policy using leave-one-out baselines with per-environment advantage normalization, then synchronizes updated weights back to vLLM via callbacks.}
    \label{fig:system}
\end{figure}

The postprocessing pipeline requires high-throughput episode generation to collect enough training data. We built a training system comprising four subsystems (Figure~\ref{fig:system}).

\subsubsection{vLLM Asynchronous Engine}

The standard vLLM \citep{kwon2023vllm} integration in TRL \citep{vonwerra2022trl} submits all samples in a single batch and blocks until the slowest completes. This triggers the heterogeneous inference bottleneck (Challenge~\ref{sec:challenge-inference}): workers generating short responses sit idle while others deliberate.

Our integration sends each request individually through an API server, relying on vLLM's continuous batching to aggregate requests internally while returning results as soon as each completes. No worker waits for another. To scale across multiple GPUs, we deploy data-parallel vLLM workers behind a coordinator that dispatches requests round-robin and returns responses to the correct caller. Weight synchronization after each training step updates all workers atomically without restarting the inference service.

\subsubsection{Rollout Provider}

The Rollout Provider maintains a pool of worker slots running episodes concurrently. Each slot encapsulates the full episode lifecycle: the action validator checks format and rule compliance during execution (Challenge~\ref{sec:challenge-missing}), and the post-episode processing pipeline computes delayed rewards with backward attribution once the episode terminates (Challenges~\ref{sec:challenge-lose} and~\ref{sec:challenge-undefined}).

Three samplers configure each new episode. The \textbf{Environment Sampler} rotates through games round-robin so that no single game dominates gradients. The \textbf{Position Sampler} cycles uniformly through roles and positions within each game (Challenge~\ref{sec:challenge-position}), so the model learns position-agnostic strategies. The \textbf{Opponent Sampler} implements curriculum learning with maintained diversity (Challenge~\ref{sec:challenge-curriculum}): weaker opponents early in training let the model learn rules and formats, while stronger and more varied opponents are unlocked at step thresholds; earlier opponents remain in the pool throughout training to preserve generalization across skill levels.

\subsubsection{Rollout Builder \& Dataloader}

Completed episodes flow into a rolling episodes bank, decoupling episode generation from batch construction (Challenge~\ref{sec:challenge-variable}). Rather than waiting for fixed-size rollouts, we accumulate episodes asynchronously and build training batches from whatever eligible steps are available.

To balance batches across multiple dimensions (Challenge~\ref{sec:challenge-balancing}), we over-collect episodes and subsample. Only steps passing the eligibility filter are considered. Steps are bucketed into reward quantiles, with coverage enforced across bins so that the model trains on successes, near-misses, and failures in proportion. Round-robin interleaving across environments prevents longer-episode games from dominating the batch.

\subsubsection{Reinforcement Learning Trainer}

Our trainer is built on TRL \citep{vonwerra2022trl} and uses a clipped policy gradient objective with the RLOO (Reinforce Leave-One-Out) baseline \citep{ahmadian2024basicsrevisitingreinforcestyle}. In standard single-turn LLM reinforcement learning, leave-one-out advantages are computed by grouping multiple completions of the \emph{same prompt}: the baseline for each completion is the mean reward of the other completions answering the identical question. In our multi-step, multi-environment setting, no two steps share the same prompt. Instead, we group all steps by \emph{environment type}: all Codenames steps form one group, all Colonel Blotto steps another, and all Iterated Prisoner's Dilemma steps a third. Each step's advantage is then computed relative to the mean reward of other steps from the same game, yielding meaningful comparisons despite prompt diversity. This per-environment normalization also prevents reward-scale differences across games from causing cross-task interference. Callbacks synchronize updated weights to the vLLM \citep{kwon2023vllm} inference workers after each training step, closing the loop without service restarts.

\subsection{Training Configuration}

We initialized from Qwen3-8B \citep{qwen3technicalreport}, a strong open-source base model that fits the 8B-parameter constraint of the Efficient track. The opponent curriculum (Challenge~\ref{sec:challenge-curriculum}) proceeded in two phases.

\paragraph{Phase 1 (Steps 0--149).} Training began exclusively against \texttt{gpt-oss-120b} \citep{openai2025gptoss120bgptoss20bmodel}, OpenAI's open-source 120B-parameter model. We configured this opponent with varied system prompts that induced different behavioral profiles (aggressive, cooperative, unpredictable, and analytical) to provide diversity within a single model family. This phase let the model learn game rules, action formats, and basic strategies without facing opponents too strong for a novice agent.

\paragraph{Phase 2 (Steps 150+).} Once the model showed competence, we introduced frontier models accessed via OpenRouter \citep{openrouter2025}. Table~\ref{tab:opponent_curriculum} shows the expanded opponent pool with sampling weights. Higher weights indicate more frequent sampling; lower weights (0.25) reserve expensive frontier models for periodic challenge while keeping training efficient. Phase 1 opponents remained in the pool throughout training to preserve generalization across skill levels.

\begin{table}[h]
\centering
\caption{Opponent models introduced at step 150 with their sampling weights. All models accessed via OpenRouter. Phase 1 opponents (\texttt{gpt-oss-120b} variants) remain active, each with weight 1.0.}
\label{tab:opponent_curriculum}
\begin{tabular}{lc}
\toprule
Model & Weight \\
\midrule
\texttt{x-ai/grok-4-fast} & 1.00 \\
\texttt{google/gemini-2.5-flash} & 1.00 \\
\texttt{google/gemini-2.5-pro} & 0.25 \\
\texttt{openai/gpt-5-mini} & 1.00 \\
\texttt{openai/gpt-5} & 0.25 \\
\texttt{qwen/qwen3-235b-a22b-thinking-2507} & 1.00 \\
\bottomrule
\end{tabular}
\end{table}

\paragraph{Hyperparameters.} We used 64 parallel workers for rollout generation with a maximum completion length of 12,288 tokens (covering both chain-of-thought reasoning and the final action). The global batch size was 768 steps. The learning rate was $1 \times 10^{-6}$ with cosine decay and warmup; the minimum learning rate was 10\% of the peak value. We disabled the Kullback-Leibler divergence penalty \citep{CoverThomas1991} (no reference model), letting the policy diverge freely from initialization. Training ran a single Proximal Policy Optimization (PPO) \citep{schulman2017ppo} epoch per batch, following \citet{kool2019reinforce}.

\subsection{Generation Parameter Tuning}

Post-training analysis showed that win rates could be improved substantially by tuning generation parameters (temperature, top-$p$, top-$k$, min-$p$) per environment. We ran a hyperparameter sweep of 300 games per environment for each parameter combination against a mixed pool of opponents. Table~\ref{tab:param_search} shows the configurations evaluated and their win rates.

\begin{table}[h]
\centering
\caption{Generation parameter search results. Win rates (\%) are shown both including draws (W) and excluding draws (W$^*$). \textbf{Bold} indicates best performance per column. $^\star$ marks the configuration selected for Codenames; $^\dagger$ marks the configuration selected for Three-Player Iterated Prisoner's Dilemma; $^\ddagger$ marks the configuration selected for Colonel Blotto.}
\label{tab:param_search}
\small
\begin{tabular}{ccccccccc}
\toprule
& & & \multicolumn{2}{c}{Codenames} & \multicolumn{2}{c}{Col. Blotto} & \multicolumn{2}{c}{Iterated Prisoner's Dilemma} \\
\cmidrule(lr){4-5} \cmidrule(lr){6-7} \cmidrule(lr){8-9}
Temp & Top-$p$ & Top-$k$ / Min-$p$ & W & W$^*$ & W & W$^*$ & W & W$^*$ \\
\midrule
1.0$^\ddagger$ & 1.0$^\ddagger$ & n/a$^\ddagger$ & 47 & 52 & \textbf{53} & 68 & 52 & 63 \\
1.0 & 0.95 & 80 / 0.05 & 38 & 39 & 44 & 58 & 51 & 61 \\
0.9 & 0.92 & 60 / n/a & 39 & 43 & 47 & 62 & 48 & 61 \\
0.7$^\star$ & 0.80$^\star$ & 30 / 0.05$^\star$ & 47 & \textbf{54} & 44 & 59 & 54 & 65 \\
0.8 & 1.0 & $-1$ / 0.15 & 29 & 34 & 52 & 67 & 55 & 64 \\
0.6$^\dagger$ & 0.85$^\dagger$ & 30 / n/a$^\dagger$ & \textbf{49} & 51 & 48 & 62 & 60 & \textbf{74} \\
0.35 & 0.70 & 20 / 0.03 & 40 & 42 & 51 & \textbf{69} & \textbf{61} & 70 \\
\bottomrule
\end{tabular}
\end{table}

No single configuration dominated across all environments, confirming the need for environment-specific tuning. Based on these results:
\begin{itemize}
    \item \textbf{Codenames:} We selected$^\star$ temp=0.70, top-$p$=0.80, top-$k$=30, min-$p$=0.05 which achieved the best win rate excluding draws (54\%).
    \item \textbf{Colonel Blotto:} We selected$^\ddagger$ the least restrictive settings (temp=1.0, top-$p$=1.0, no truncation) which achieved the best win rate including draws (53\%) and the second best win rate excluding draws (68\%).
    \item \textbf{Three-Player Iterated Prisoner's Dilemma:} We selected$^\dagger$ temp=0.60, top-$p$=0.85, top-$k$=30 which achieved the best win rate excluding draws (74\%).
\end{itemize}

Table~\ref{tab:generation_params} summarizes the final per-environment generation parameters used for evaluation. Default vLLM values are: temperature $= 1.0$, top-$p$ $= 1.0$, top-$k$ $= -1$ (consider all tokens), and min-$p$ $= 0.0$.

\begin{table}[h]
\centering
\caption{Per-environment generation parameters. Values marked with $^\dagger$ indicate vLLM defaults.}
\label{tab:generation_params}
\begin{tabular}{lcccc}
\toprule
Environment & Temperature & Top-$p$ & Top-$k$ & Min-$p$ \\
\midrule
Codenames & 0.70 & 0.80 & 30 & 0.05 \\
Colonel Blotto & 1.0$^\dagger$ & 1.0$^\dagger$ & $-1^\dagger$ & 0.0$^\dagger$ \\
Three-Player Iterated Prisoner's Dilemma & 0.60 & 0.85 & 30 & 0.0$^\dagger$ \\
\bottomrule
\end{tabular}
\end{table}

\section{Results}
\label{sec:results}

We report results from MindGames Arena \citep{mindgames2025}, where top-qualifying teams competed head-to-head across the three game environments described in Section~\ref{sec:competition}. Our model placed first in both tracks during Stage 1 qualification and held this position through the final Stage 2 evaluation, finishing \textbf{first in both tracks} (Open and Efficient) with the same 8-billion-parameter model.

\begin{table}[h]
\centering
\caption{Open Track final results (Stage 2). Best submission per team, ranked by TrueSkill \citep{herbrich2006trueskill}.}
\label{tab:open_results}
\begin{tabular}{clccc}
\toprule
Rank & Team & TrueSkill & Games & Win Rate \\
\midrule
\textbf{1} & \textbf{In2AI (ours)} & \textbf{38.0}$_{\pm 1.8}$ & \textbf{116} & \textbf{81.0\%} \\
2 & RLGaming & 37.1$_{\pm 1.1}$ & 291 & 73.5\% \\
3 & Odyssean & 34.2$_{\pm 1.4}$ & 177 & 72.3\% \\
4 & PsychSkull & 31.3$_{\pm 1.4}$ & 121 & 62.8\% \\
5 & Corleone & 28.6$_{\pm 1.3}$ & 117 & 49.6\% \\
\bottomrule
\end{tabular}
\end{table}

\begin{table}[h]
\centering
\caption{Efficient Track ($\leq$8B) final results (Stage 2), ranked by TrueSkill \citep{herbrich2006trueskill}.}
\label{tab:efficient_results}
\begin{tabular}{clccc}
\toprule
Rank & Team & TrueSkill & Games & Win Rate \\
\midrule
\textbf{1} & \textbf{In2AI (ours)} & \textbf{34.2}$_{\pm 1.3}$ & \textbf{362} & \textbf{87.0\%} \\
2 & STARS & 26.8$_{\pm 1.1}$ & 337 & 36.2\% \\
3 & RLGaming & 25.8$_{\pm 1.1}$ & 569 & 28.5\% \\
4 & Corleone & 24.4$_{\pm 1.4}$ & 367 & 44.1\% \\
5 & Odyssean & 16.6$_{\pm 1.4}$ & 275 & 10.9\% \\
\bottomrule
\end{tabular}
\end{table}

Table~\ref{tab:open_results} shows the Open Track results, where teams could use any model without restrictions on size or cost. Despite competing against submissions built on frontier models such as GPT-5 \citep{openai2025gpt5}, our 8B-parameter model achieved the highest TrueSkill~\citep{herbrich2006trueskill} rating and win rate. Table~\ref{tab:efficient_results} presents the Efficient Track results, restricted to models with at most 8 billion parameters; here, our model led by over 7 TrueSkill points.

The win rates across both tracks indicate that careful reward attribution and eligibility gating, paired with balanced training, allow a smaller model to compete with and outperform substantially larger systems across varied opponents and game types within the MindGames Arena \citep{mindgames2025} evaluation framework.

\section{Conclusion}
\label{sec:conclusion}

We presented an approach for training agentic large language models centered on \emph{delayed per-step reward attribution with eligibility gating}. Standard reinforcement learning assumes each step can receive an immediate, well-defined reward; this assumption fails in agentic settings where action quality depends on future events beyond any single agent's control.

We identified eight challenges in multi-agent strategic interaction, organized into three themes: temporal entanglement (Sections~\ref{sec:challenge-lose}--\ref{sec:challenge-undefined}), structural asymmetry (Sections~\ref{sec:challenge-position}--\ref{sec:challenge-missing}), and training logistics (Sections~\ref{sec:challenge-variable}--\ref{sec:challenge-inference}). We addressed them through a post-episode processing pipeline that delays reward computation until sufficient information is available, attributes rewards backward according to task-specific semantics, and gates steps with undefined outcomes from training. Paired with high-throughput asynchronous infrastructure (continuous-batching inference, curriculum-based opponent sampling, and balanced batch construction), the system trains efficiently at scale.

A single 8-billion-parameter open-source model trained with this approach placed first in both the Open (unrestricted) and Efficient ($\leq$8B) tracks of MindGames Arena~\citep{mindgames2025}, outperforming teams that used substantially larger proprietary systems including GPT-5~\citep{openai2025gpt5} in head-to-head competition.

\paragraph{Limitations and Future Work.}
Our approach requires environment-specific implementations of the Action Validator, Players Builder, Steps Filter, and Reward Assigner; while the underlying ideas generalize, adapting to new settings requires domain expertise. Broader validation across agentic tasks beyond the three game environments studied here is an important next step.

\paragraph{Broader Impact.}
In our experience, the engineering challenges outlined in Section~\ref{sec:challenges} were a bigger bottleneck than model size; careful reward attribution mattered more than scaling up parameters. This suggests that \textbf{systems-level engineering, not model scale, determines success in agentic RL}. We expect these ideas to extend to other settings where action outcomes depend on future events, such as multi-turn dialogues, collaborative problem-solving, and agentic code generation.

\paragraph{Reproducibility.}
Our trained model weights are publicly available at \url{https://huggingface.co/AlekseyKorshuk/mindgames-in2ai-submission}. The training code and evaluation scripts are available at \url{https://github.com/AlekseyKorshuk/mindgames-in2ai}.

\section*{Acknowledgements}

We thank the MindGames Arena~\citep{mindgames2025} organizers for creating the evaluation framework and for hosting the competition at NeurIPS 2025. We are grateful to the developers of TextArena~\citep{guertler2025textarena} for the game environments and infrastructure. We also thank the vLLM~\citep{kwon2023vllm} and TRL~\citep{vonwerra2022trl} teams, whose inference engine and RL framework form the backbone of our training system.

\bibliographystyle{unsrtnat}
\bibliography{references}

\appendix

\section{Episode-Level Reward Computation (Players Builder)}
\label{app:players-builder}

As discussed in Section~\ref{sec:players-builder}, the Players Builder computes meaningful episode-level rewards from the raw game state. This section provides the exact formulas used for each environment.

\subsection{Colonel Blotto}

In Colonel Blotto, the game consists of $R$ rounds (typically $R = 9$), and a player wins by securing a majority of rounds. Let $W^{(p)}$ denote the number of rounds won by player $p$, and let $M = \lfloor R/2 \rfloor + 1$ be the majority threshold (rounds needed to win the match). The episode reward is:

\begin{equation}
r_{\text{episode}}^{(p)} = \frac{W^{(p)}}{M}
\end{equation}

For $R = 9$, we have $M = 5$, so:
\begin{itemize}
    \item Winning 5 rounds yields $r_{\text{episode}} = 1.0$
    \item Winning 3 rounds yields $r_{\text{episode}} = 0.6$
    \item Winning 0 rounds yields $r_{\text{episode}} = 0.0$
\end{itemize}

This normalization makes rewards reflect progress toward victory rather than the binary win/loss outcome alone.

\subsection{Codenames}

In Codenames, each team has a target number of words to guess: $G_R = 9$ for Red and $G_B = 8$ for Blue. Let $C^{(p)}$ denote the number of correct team words guessed by team $p$'s operative, and let $G^{(p)}$ be that team's goal. The episode reward is:

\begin{equation}
r_{\text{episode}}^{(p)} = \frac{C^{(p)}}{G^{(p)}}
\end{equation}

For a Red team player:
\begin{itemize}
    \item Guessing all 9 words yields $r_{\text{episode}} = 1.0$
    \item Guessing 5 words yields $r_{\text{episode}} \approx 0.56$
    \item Guessing 0 words yields $r_{\text{episode}} = 0.0$
\end{itemize}

This reward is computed \emph{per team}: both the spymaster and operative on the same team receive the same episode-level reward, since they share responsibility for collective performance.

\subsection{Three-Player Iterated Prisoner's Dilemma}

In the three-player IPD, each player interacts with two opponents over $N$ rounds (typically $N = 5$). The payoff matrix follows the standard structure with values $R$ (reward for mutual cooperation), $T$ (temptation to defect), $S$ (sucker's payoff), and $P$ (punishment for mutual defection). Let $S^{(p)}$ denote player $p$'s cumulative score across all rounds and interactions.

The maximum possible score per round is $2 \cdot \max(R, T, S, P)$ (one interaction with each opponent). The episode reward is normalized by the theoretical maximum:

\begin{equation}
r_{\text{episode}}^{(p)} = \frac{S^{(p)}}{2 \cdot \max(R, T, S, P) \cdot N}
\end{equation}

With standard payoffs ($R=3, T=5, S=0, P=1$) and $N=5$ rounds:
\begin{itemize}
    \item Maximum score is $2 \times 5 \times 5 = 50$ (defecting against two always-cooperating opponents)
    \item Mutual cooperation yields $2 \times 3 \times 5 = 30$, so $r_{\text{episode}} = 0.6$
    \item Mutual defection yields $2 \times 1 \times 5 = 10$, so $r_{\text{episode}} = 0.2$
\end{itemize}

\subsection{Outcome Determination}

Beyond reward computation, the Players Builder determines categorical outcomes (win/loss/draw/invalid) using the following priority:

\begin{enumerate}
    \item \textbf{Invalid action:} If any player made an invalid action, that player receives \texttt{invalid\_action} outcome; others receive \texttt{other\_player\_invalid\_action}.
    \item \textbf{Parsing failure:} Same logic applies for parsing failures.
    \item \textbf{Game-specific termination:} Assassin hits (Codenames), subset clues (Codenames), or other rule violations.
    \item \textbf{Score comparison:} Higher score wins; equal scores result in draw.
\end{enumerate}

This structured outcome determination feeds the Steps Filter and Reward Assigner with the information they need for step eligibility and reward attribution decisions.

\section{Per-Step Reward Attribution}
\label{app:rewards}

This section details the per-step reward assignment and step filtering logic for each game environment. All rewards are clipped to $[-1, 1]$ with strict floors for invalid actions ($-0.75$) and parsing failures ($-1.0$).

\subsection{Codenames}

Codenames presents unique challenges because spymaster clues must be evaluated based on operative guesses, creating inter-step dependencies.

\subsubsection{Action Validation}

\begin{itemize}
 \item \textbf{Spymasters} (players 0, 2): Must produce \texttt{[word N]} where \texttt{word} is a single alphabetic token and \texttt{N} is a positive integer.
 \item \textbf{Operatives} (players 1, 3): Must produce \texttt{[word]} where \texttt{word} is a single token (or ``pass'').
 \item \textbf{Substring rule:} Clues that are substrings of (or contain) any board word are marked as violations, receiving penalty $-0.5$.
\end{itemize}

\subsubsection{Clue Reward (Spymaster)}

For a clue with declared count $N$ when $L$ team words remain unguessed, let $N_{\text{eff}} = \min(N, L)$. After the operative makes $G$ correct guesses (out of the $N_{\text{eff}}$ attempts within scope):

\begin{equation}
r_{\text{clue}} = \underbrace{\frac{N_{\text{eff}}}{L}}_{\text{ambition}} \times \underbrace{\frac{G}{N_{\text{eff}}}}_{\text{efficiency}} + \Delta_{\text{blame}}
\end{equation}

where $\Delta_{\text{blame}}$ adds fixed penalties if the turn resulted in guessing an opponent word ($-0.25$) or the assassin ($-0.50$). Special cases:
\begin{itemize}
 \item $N < 1$: reward $= -0.6$ (invalid clue count)
 \item $N > L$ (overshooting): linear penalty in $[-0.5, -0.25]$ based on overshoot magnitude
 \item $G = 0$ (no correct guesses): $r_{\text{clue}} = -0.25 \times (N_{\text{eff}}/L)$
\end{itemize}

\subsubsection{Guess Reward (Operative)}

For guess $i$ within the scope of a clue ($i \leq N_{\text{eff}}$):

\begin{center}
\begin{tabular}{lc}
\toprule
Condition & Reward \\
\midrule
Correct (own team word) & $\bar{r}$ (group average) \\
Neutral word & $-0.40$ \\
Duplicate guess & $-0.50$ \\
Off-board word & $-0.55$ \\
Opponent word & $-0.60$ \\
Assassin & $-0.70$ \\
Early pass (within $N$) & $-0.60$ \\
\bottomrule
\end{tabular}
\end{center}

For the $N+1$th guess (bonus attempt), passing is rewarded ($+0.5$), while continuing risks penalties. Correct guesses receive the \emph{group average reward} $\bar{r}$, computed over all guesses within scope, so that operatives share credit and blame for collective performance.

\subsubsection{Outcome Multipliers}

Positive step rewards are scaled by episode outcome: win $\times 1.0$, draw $\times 0.6$, loss $\times 0.2$.

\subsubsection{Step Filtering}

\begin{itemize}
 \item Steps from non-trainable players are excluded.
    \item Clue steps with \texttt{condition = correct} but \texttt{len(guesses) = 0} are excluded (no signal available).
 \item Invalid/parsing-failed steps remain eligible (to train format compliance).
\end{itemize}

\subsection{Colonel Blotto}

Colonel Blotto involves multiple rounds of resource allocation with immediate per-round feedback.

\subsubsection{Round-Level Reward}

Each round produces a winner (or tie). For player $p$ in round $r$:

\begin{equation}
r_{\text{round}}^{(p)} = \begin{cases}
1.0 & \text{if } p \text{ wins round } r \\
0.6 & \text{if tie} \\
0.0 & \text{if } p \text{ loses round } r
\end{cases}
\end{equation}

\subsubsection{Match-Level Shaping}

The match outcome (determined by who wins more rounds) modulates step rewards:

\begin{equation}
r_{\text{step}}^{(p)} = \kappa \cdot m^{(p)} \cdot r_{\text{round}}^{(p)}
\end{equation}

where $\kappa = 1.0$ and the match multiplier $m^{(p)}$ is:
\begin{center}
\begin{tabular}{lc}
\toprule
Match Outcome & $m^{(p)}$ \\
\midrule
Win & $1.0$ \\
Draw & $0.6$ \\
Loss & $0.2$ \\
\bottomrule
\end{tabular}
\end{center}

Winning individual rounds therefore matters more when they contribute to match victory.

\subsubsection{Step Filtering}

\begin{itemize}
 \item All trainable player steps are initially eligible.
 \item If the opponent made an invalid move (ending the match early), the last step of the trainable player is excluded if it occurred after the invalid action, since no learning signal is available for that incomplete round.
\end{itemize}

\subsection{Three-Player Iterated Prisoner's Dilemma}

Iterated Prisoner's Dilemma combines conversation rounds with binary decisions, requiring group-based reward attribution.

\subsubsection{Step Grouping}

Each Iterated Prisoner's Dilemma round consists of $C$ conversation steps followed by 1 decision step, forming a group of size $C+1$. Rewards are attributed to the entire group based on the decision outcome.

\subsubsection{Round Gain Normalization}

Let $\Delta_r^{(p)}$ be player $p$'s score increment in round $r$. The group reward is:

\begin{equation}
r_{\text{group}}^{(p)} = \frac{\Delta_r^{(p)}}{2 \cdot \max(R, T, S, P)}
\end{equation}

where $R, T, S, P$ are the standard Iterated Prisoner's Dilemma payoff matrix values (Reward, Temptation, Sucker, Punishment). The denominator normalizes by the maximum possible per-interaction gain.

\subsubsection{Reward Distribution}

The group reward is assigned equally to all valid steps in the group (conversation + decision). If the decision step is invalid or parsing-failed:
\begin{itemize}
 \item The decision step receives its standard penalty ($-0.75$ or $-1.0$).
 \item All conversation steps in that group receive reward $0.0$ (no propagation of positive signal to actions that led to an invalid decision).
\end{itemize}

\subsubsection{Outcome Multipliers}

Positive step rewards are scaled by episode outcome: win $\times 1.0$, draw $\times 0.6$, loss $\times 0.2$.

\subsubsection{Step Filtering}

\begin{itemize}
 \item All trainable player steps are initially eligible.
 \item \textbf{Incomplete groups:} If a round ends mid-conversation (before the decision step), all non-error conversation steps in that incomplete group are marked ineligible because they have no associated outcome.
 \item \textbf{Invalid decision propagation:} If the decision step is invalid, preceding conversation steps in that group are marked ineligible (no positive signal to propagate), but error steps remain eligible for training.
\end{itemize}

\subsection{Summary}

The key principles across all environments are:

\begin{enumerate}
 \item \textbf{Strict error floors:} Parsing failures ($-1.0$) and invalid actions ($-0.75$) always dominate shaped rewards, so format compliance is learned quickly.
 \item \textbf{Delayed attribution:} Step rewards may depend on future events (guesses after clues, decisions after conversations), computed only after episode completion.
 \item \textbf{Signal-based filtering:} Steps without valid dependent information are excluded rather than trained with arbitrary rewards.
 \item \textbf{Outcome shaping:} Episode-level outcomes modulate step-level rewards, connecting local decisions to global success.
\end{enumerate}

\end{document}